\documentclass[runningheads]{llncs}
\usepackage[T1]{fontenc}
\usepackage{amsmath}
\usepackage{array}
\usepackage{bbm}

\usepackage{float}
\usepackage{amsbsy}
\usepackage{amstext}
\usepackage{amssymb}
\usepackage{graphicx}
\usepackage{svg}
\usepackage{dsfont}
\usepackage[T1]{fontenc}
\usepackage[font=small,labelfont=bf,tableposition=top]{caption}

\begin{document}
\title{LaTiM: Longitudinal representation learning in continuous-time models to predict disease progression}

\titlerunning{Longitudinal representation learning in continuous-time models}

\author{Rachid Zeghlache\inst{1,2}
\and
Pierre-Henri Conze\inst{1,3} 
\and
Mostafa El Habib Daho \inst{1,2}
\and
Yihao Li \inst{1,2}
\and
Hugo Le Boité \inst{5}
\and
Ramin Tadayoni\inst{5} 
\and
Pascal Massin\inst{5} 
\and
Béatrice Cochener\inst{1,2,4} 
\and
Alireza Rezaei \inst{1,2} 
\and
Ikram Brahim \inst{1,2} 
\and 
Gwenolé Quellec\inst{1} 
\and
Mathieu Lamard\inst{1,2} 
}


\authorrunning{R. Zeghlache et al. }
%


\institute{
LaTIM UMR 1101, Inserm, Brest, France \and
University of Western Brittany, Brest, France \and
IMT Atlantique, Brest, France
\and
Ophtalmology Department, CHRU Brest, Brest, France \and
Lariboisière Hospital, AP-HP, Paris, France
}
\maketitle
\begin{abstract}

This work proposes a novel framework for analyzing disease progression using time-aware neural ordinary differential equations (NODE). We introduce a "time-aware head" in a framework trained through self-supervised learning (SSL) to leverage temporal information in latent space for data augmentation. This approach effectively integrates NODEs with SSL, offering significant performance improvements compared to traditional methods that lack explicit temporal integration. We demonstrate the effectiveness of our strategy for diabetic retinopathy progression prediction using the OPHDIAT database. Compared to the baseline, all NODE architectures achieve statistically significant improvements in area under the ROC curve (AUC) and Kappa metrics, highlighting the efficacy of pre-training with SSL-inspired approaches. Additionally, our framework promotes stable training for NODEs, a commonly encountered challenge in time-aware modeling.

\end{abstract}

\section{Introduction}

Deep learning has embraced self-supervised learning (SSL) for representation learning in downstream tasks. Existing SSL methods often rely on contrastive learning \cite{SimCLR} or hand-crafted pretext tasks \cite{zhang2016colorful}. The latter leverages inherent data properties to automatically generate supervisory signals without manual annotations. While effective, hand-crafted tasks require domain-specific knowledge. Recently, longitudinal SSL approaches have emerged for disease progression analysis, aiming to capture disease evolution at patient or population levels \cite{Vernhet2021,ren2022local,Zhao2021,Couronné,Ren2022,ouyang2023lsor,Holland2022MetadataenhancedCL}. Longitudinal self-supervised learning (LSSL) was initially introduced in the context of disease progression as a pretext task \cite{Rivail2019} involving a Siamese-like model. The model takes as input a consecutive pair of images and predicts the difference in time between the two examinations. Since then, more sophisticated LSSL algorithms have been proposed. The framework in \cite{Zhao2021} attempted to theorize the notion of longitudinal pretext task with the purpose of learning the disease progression. LSSL was embedded in an auto-encoder (AE), taking two consecutive longitudinal scans as inputs. A cosine alignment term was added to the classic reconstruction loss to force the topology of the latent space to change in the direction of longitudinal changes. A Siamese-like architecture was developed in \cite{Rivail2019,kim2023learning} to compare longitudinal imaging data with deep learning. The strengths of this approach were to avoid any registration requirements, leverage population-level data to capture time-irreversible changes that are shared across individuals and offer the ability to visualize individual-level changes. The previously mentioned methods primarily focus on learning an adequate feature extractor for disease progression analysis. While these approaches offer valuable contributions, we claim that incorporating a time-aware component during the learning process can lead to more accurate disease progression models. 

\begin{figure}[t]
\centering
\includegraphics[width=0.89\textwidth]{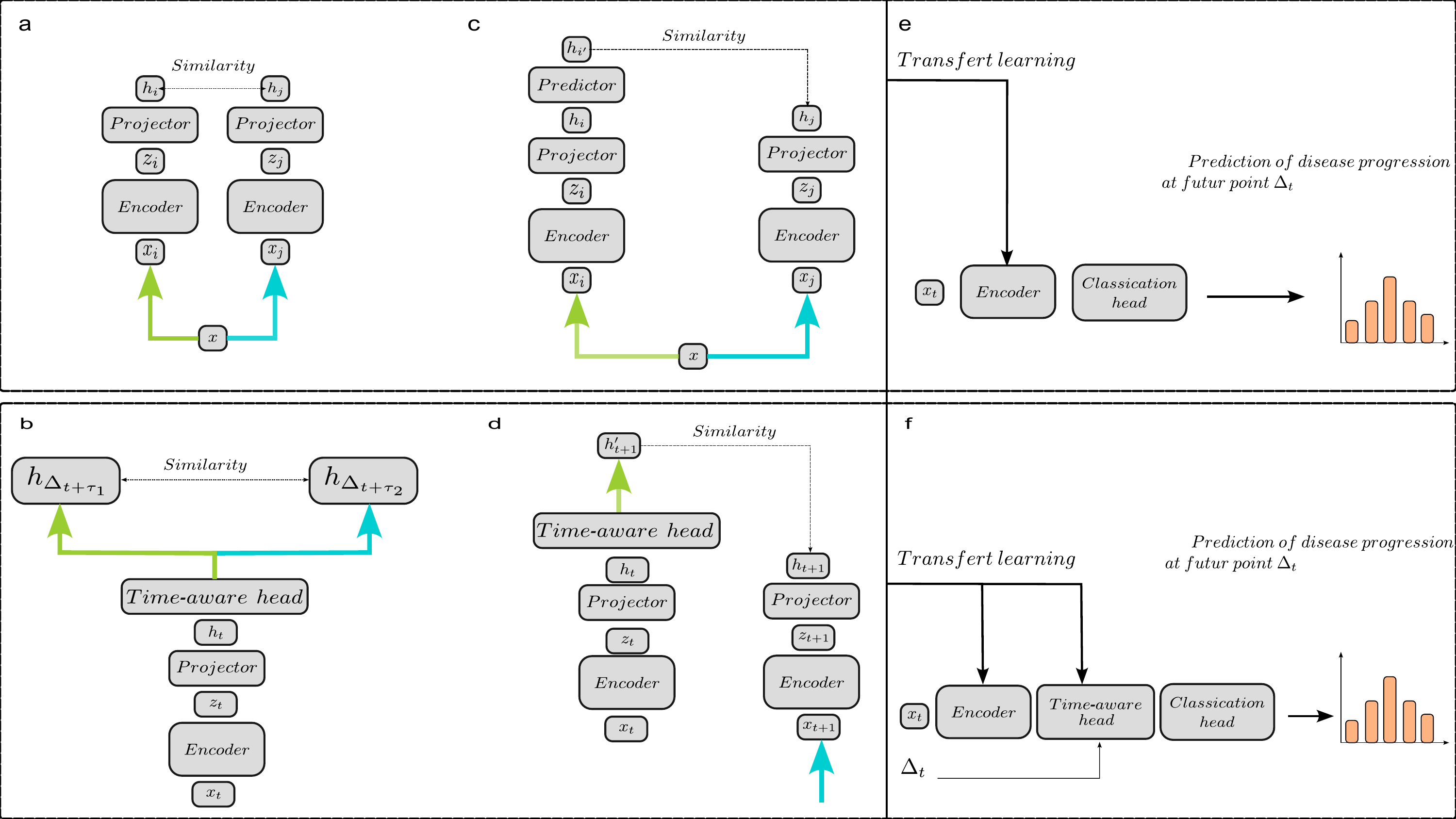}
\caption{a), c) refer to SimCLR \cite{SimCLR} and BYOL \cite{grill2020bootstrap}, and e) refer to the standard paradigm for predicting disease progression at a certain point in the future, using a single image. b), d), and f) adapt the common SSL paradigm and supervised classification to disease progression with the introduction of a time-aware head.} 
\label{fig:overview}
\end{figure}
Neural ordinary differential equations (NODEs) \cite{NeuralODE} offer a promising approach for modeling time-dependent processes like disease progression. NODEs define the relationship between input and output through the solution of an ordinary differential equation (ODE). Moreover, this allows to effectively handle irregular time series data, a common feature in disease progression analysis \cite{Yulia}. However, current disease progression models often lack a dedicated time-aware component. We propose to integrate a time-aware head based on a NODE architecture between the encoding and classification stages. This approach aims to condition the predictions with time, towards more accurate models \cite{Lachinov2023,zeghlache2023lmt,zeghlache-prime}.

Recent works support the use of NODEs for disease progression. Studies have employed NODEs to model COVID-19 progression \cite{NEURIPS2021} and predict Alzheimer's disease progression \cite{Lachinov2023}. This work investigates the hypothesis that incorporating a time-aware head based on NODE can outperform traditional classification heads for disease progression tasks \cite{Bora2021}. We claim that this approach can lead to a more informative latent space representation, facilitating more advanced disease progression analysis. To establish our proposed framework, we define and train a set of similarity criteria specifically tailored to the learning paradigm. These criteria, referred to as \textit{temporal evolution} (TE), \textit{temporal consistency} (TC), and \textit{disease progression alignment} (DPA), are critical to guide the learning process.

This paper presents the following key contributions:

\begin{enumerate}
    \item We propose a novel framework for pre-training time-aware models to address disease progression downstream tasks, specifically focusing on diabetic retinopathy progression.
    \item We explore the application of SSL to NODEs for disease progression analysis.
    \item We bridge the gap between SSL and time-aware models by introducing novel temporal augmentations.
\end{enumerate}

To the best of our knowledge, this is the first attempt to leverage SSL to learn improved weights specifically for NODE-based models in the context of disease progression analysis using longitudinal medical images.

\section{Methods}

This section describes the methodology developed for disease progression analysis. We start by briefly describing two important frameworks commonly used for pre-training. Our different hypotheses will be used on top of these frameworks. To harmonize the notation, let $f$, $g$, $u$ be encoder, projection head, and time-aware head, respectively. The core novelty lies in employing NODEs within the BYOL/SimCLR framework. With the proposition of two straightforward augmentation schemes to pre-train a time-aware head using the most popular SSL paradigms: SimCLR \cite{SimCLR} and BYOL \cite{grill2020bootstrap}. Let $\mathcal{V}$ be the set of consecutive patient-specific image pairs from the collection of all CFP images. $\mathcal{V}$ contains all $(\mathbf{x}_{t_{i}}, \mathbf{x}_{t_{i+1}})$ that are from the same patient where $\mathbf{x}_{t_{i}}$ is scanned before $\mathbf{x}_{t_{i+1}}$ and $i \in [0,m-2]$ with $m$ the number of visits for a given eye. For a given patient, the scans are irregularly sampled due to clinical constraints ($\Delta_{t_{i}} \neq  \Delta_{t_{i+1}}$) with $\Delta_{t_{i}} = t_{i+1}-t_{i}$ the time difference between two consecutive exams and we denote  $S_{t_{i}}$ the severity grade at time $t_{i}$. Time-aware models are deep neural networks that take as inputs both time and embedding representation. We note $h'_{t_{i+1}} = u(h(t_i), t_i, t_{i+1}, \theta)$ where $u$ denotes a given deep neural network, and $\theta$ its trainable parameters and $h_{t_{i}}$ some latent vector at time $t_{i}$ and $h'_{t_{i+1}}$ the latent representation produced by $u$ at time $t_{i+1}$. In our experiments, Neural ODE (NODE) based methods will be followed.

\begin{figure}[h!]
\centering
\includegraphics[width=\textwidth]{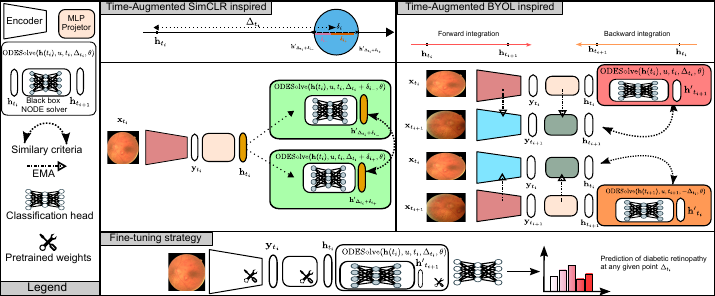}

\caption{Proposed augmentation techniques that we employed to mimic the popular SSL paradigms (SimCLR \cite{SimCLR} on the left, BYOL \cite{grill2020bootstrap} on the right) with neural ODE.} 

\label{fig:BYOL_SimCLR}
\end{figure}

\noindent \textbf{Neural Ordinary Differential Equations (NODEs)} approximate unknown ordinary differential equations by a neural network \cite{NeuralODE} that parameterizes the continuous dynamics of hidden units $h\in \mathbb{R}^n$ over time with $\mathbf{t}\in \mathbb{R}$. NODEs are able to model the instantaneous rate of change of $h$ with respect to $\mathbf{t}$ using a neural network $u$ with parameters $\theta$.

\begin{equation} 
\label{equ:neural_odes}
\lim_{\epsilon\rightarrow0}\frac{h_{t+\epsilon}-h_t}
{\epsilon}=\frac{dh}{dt}=u(t,h,\theta)
\end{equation} \vspace{-0.1cm} \\

\noindent The analytical solution of Eq.\ref{equ:neural_odes} is given by:
 
\begin{equation} 
h_{t_{i+1}} = h_{t_{i}} + \int_{t_{i}}^{t_{i+1}}u(t,h,\theta)\mathrm{d}t =\textrm{ODESolve}(h(t_i), u, t_i, t_{i+1}, \theta)\label{equ:solveode}
\end{equation}

\noindent where $[t_{i}, t_{i+1}]$ represents the time horizon for solving the ODE, $u$ being a neural network, and $\theta$ is the trainable parameters of $u$. By using a black-box ODE solver introduced in \cite{NeuralODE}, we can solve the initial value problem (IVP) and calculate the hidden state at any desired time using Eq.\ref{equ:solveode}. We can differentiate the solutions of the ODE solver with respect to the parameters $\theta$, the initial state $\mathbf{h}_{t_{i}}$ at initial time $t_{i}$, and the solution at time $t$. This can be achieved by using the adjoint sensitivity method \cite{NeuralODE}. Through the latent representation of a given image, we define an IVP that aims to solve the ODE from $t_{i}$ to a terminal time $t_{i+1}$:
\begin{equation}
    \dot h(t) = u(h(t), t, \theta), \textrm{with the initial value} \:h(t_{i}) = h_{{t_{i}}}
    \label{eq:IVP}
\end{equation}

Following the same way, we can define a final value problem (FVP). This is possible due to the fact that neural ODE are invertible and learn homeomorphic transformation \cite{NeuralODE}. Through the latent representation of a given image, we define a FVP that aims to solve the ODE from $t_{i+1}$ to a reverse time $t_{i}$:

\begin{equation}
    \dot h(t) = u(h(t), -t, \theta), \textrm{with the final value} \:h(t_{i+1}) = h_{{t_{i+1}}}
    \label{eq:FVP}
\end{equation}

\subsection{SimCLR-based approach: \textit{disease progression alignment}}

During training, we sample pairs from $\mathcal{V}$ and encode the first element of the pair $\mathbf{y}_{t_{i}}=f(\mathbf{x_{t_{i}}})$ using the encoder $f$. Then, using the projector $g$, we obtain the representation of $h_{t_{i+1}} = g(\mathbf{y}_{t_{i}})$. To transpose the SimCLR (Fig.\ref{fig:overview}.a) learning paradigm by defining close time in the patient disease trajectory such that the transformation plays the role of augmentation for the time-aware model:  $h'_{\Delta_{t_{i}} + \delta_{i}}=\textrm{ODESolve}(h(t_{i}), u, t_{i},\Delta_{t_{i}}+\delta_{i}, \theta)$. The value of $\delta_{i}$ is chosen to align with the known progression of the disease. For a given pair of image $(\mathbf{x}_{t_{i}},\mathbf{x}_{t_{i+1}})$, we first calculate the rate of variation such that $r_{i}=\frac{S_{t_{i+1}}-S_{t_{i}}}{{t_{i+1}}-{t_{i}}}$ . By using $r_{i}$, we define an interval $\delta_{i}$ where the disease rate of progression is the smallest on a period of one month, with $r_{i}\cdot \frac{12}{365}$. Using  $\delta_{i}$ we now define $\delta_{i_{+}}$ and $\delta_{i_{-}}$ two sub-intermediate time, which will play the role of time augmentation for the neural NODE when solving the ODE. We choose $\delta_{i_{+}} \sim \text{Unif}(0,\delta_{i})$ (\text{Unif} denote the uniform distribution) and $\delta_{i_{-}} = \delta_{i} - \delta_{i_{+}}$ so that the maximum time distance between the two sampled times is at most $\delta_{i}$. A schematic representation of the operations is illustrated in Fig.\ref{fig:BYOL_SimCLR} (left). We follow the standard SimCLR loss where $\tau$ and $sim$ stand for the temperature and the cosine similarity, respectively. We can rewrite the objective loss  for one pair in a batch:

\begin{centering}
\begin{align}
h'_{\Delta_{t_{i}} + \delta_{i_{+}}} &= \textrm{ODESolve}(h(t_{i}), u, t_{i}, \Delta_{t_{i}}+\delta_{i_{+}}, \theta),\quad \\
h'_{\Delta_{t_{i}} - \delta_{i_{-}}} &= \textrm{ODESolve}(h(t_{i}), u, t_{i}, \Delta_{t_{i}}-\delta_{i_{-}}, \theta),\quad  \\
\mathcal{L} &= - \log\frac{\exp(\text{sim}(h'^{i}_{\Delta_{t_{i}} + \delta_{i_{+}}}, h'^{i}_{\Delta_{t_{i}} + \delta_{i_{-}}}) / \tau)}{\sum_{k=1}^{2N} \mathbbm{1}_{[k \neq i]} \exp(\text{sim}(h'^{i}_{\Delta_{t_{i}} + \delta_{i_{+}}}, h'^{k}_{\Delta_{t_{i}} + \delta_{i_{+}}} / \tau)}
\end{align}
\end{centering}

\subsection{BYOL-based approach: \textit{temporal evolution} and \textit{temporal consistency}}\label{secsub:BYOLinspired}

Similar to BYOL, we leverage two neural networks, denoted to as online (parameterized by $\mu$) and target (parameterized by $\xi$), that interact and guide each other's learning. The target network shares the same architecture as the online network but uses exponentially moving average weights  $\xi \leftarrow \alpha \xi + (1-\alpha) \mu$ with $\alpha$ controlling the moving average strength. During training, a pair of data ($\mathbf{x}_{t_{i}}$,$\mathbf{x}_{t_{i+1}}$) are sampled from the data distribution $\mathcal{V}$. These are encoded by the online and target networks' shared encoder $f$ into representations $\mathbf{y}_{t_{i}}$ and $\mathbf{y}_{t_{i+1}}$, respectively. Subsequently, projectors $g_{\mu}$ and $g_\xi$ project them into latent representations $h_{t_{i}}$ and $h_{t_{i+1}}$. The online network leverages the initial latent code $h_{t_{i}}$ to predict the future latent code $h'_{t_{i+1}}$ using the IVP of the neural ODE (Eq.\ref{eq:IVP}). Both predicted and target representations are L2-normalized. The BYOL loss $\mathcal{L}^\text{BYOL NODE}_\mu$ is the mean squared error between the normalized prediction $h'_{t_{i+1}}$  and $h_{t_{i+1}}$. A complementary symmetric loss $\tilde{\mathcal{L}}^\text{BYOL NODE}_\mu$ is formulated using the FVP of the neural ODE, predicting $h'_{t_{i}}$ from ${h_{t_{i+1}}}$ (Eq.\ref{eq:FVP}). Fig.\ref{fig:BYOL_SimCLR} (right) visually depicts this process. The final loss function combines both BYOL losses: $$ \mathcal{L} = \mathcal{L}^\text{BYOL NODE}_\mu + \tilde{\mathcal{L}}^\text{BYOL NODE}_\mu$$

\section{Experiments and results}

\subsection{Dataset and implementation details.} We trained and evaluated our models on OPHDIAT \cite{OPHDIAT}, a vast dataset of fundus photographs (CFPs) collected from the Ophthalmology Diabetes Telemedicine network. This dataset encompasses examinations from over 101,000 patients between 2004 and 2017. Among the 763,848 interpreted CFP images, nearly 673,000 received diabetic retinopathy (DR) severity grade, while the remainder were ungradable. The patient's age ranges from 9 to 91 years. We trained the networks for 400 epochs using AdamW optimizer, OneCycleLR scheduler,learning rate of 1e-3, weight decay of 1e-4, and a batch size of 128 on a single Nvidia A6000 GPU with PyTorch. For NODEs, we utilized the Torchdiffeq library \cite{torchdiffeq} for solving ODEs, back-propagation, and the adjoint method. Our NODE architecture comprised dense layers with tanh activation and dopri5 solver. We monitored validation performance and saved the best model. We used ResNet50 \cite{he2015deep} as a backbone and a two-layer MLP projector. By leveraging consecutive image pairs, we achieved diversity without additional augmentations, fostering a robust time-aware predictor for longitudinal tasks. We evaluated the effectiveness of our framework in enhancing time-continuous models (TMCs) by comparing our approach with popular TMCs like NODE  \cite{NeuralODE}, NODE-RNN \cite{Yulia}, NODE-GRU \cite{Brouwer2019GRUODEBayesCM}, and NODE-LSTM  \cite{lechner2020learning}. We evaluated our pre-trained model by fine-tuning the time-aware and feature extractor component for 100 epochs on two tasks: predicting DR progression at 1-3 years (task 1), and next visit timing (task 2) . The area under the receiver operating characteristic curve (AUC) with three binary tasks: predicting at least non-proliferative diabetic retinopathy (NPDR), (\textbf{AUC1}), at least moderate NPDR (\textbf{AUC2}) and finally at least severe NPDR (\textbf{AUC3}) and the quadratic-weighted Kappa were used metrics. When $\delta_{i}$ is null a default disease rate of progression equivalent of 3 month is given. Further training dataset details and hyper-parameters are given in supplementary materials. 

\subsection{Results}

\noindent \textbf{Evaluation of time aware model on a fixed time interval (Task 1)}: Prediction of diabetic retinopathy progression for 1,2,3 years. Tab.\ref{tab:1_2_3_pred} compares methods for predicting diabetic retinopathy (DR) progression at 1, 2, and 3 years using AUC. A baseline model (ResNet50) achieved moderate performance (AUC 0.57-0.66). Adding a NODE layer generally decreased performance, and further incorporating recurrent layers (RNN, LSTM, GRU) yielded mixed results.  However, pre-training the NODE layer with either disease progression alignment or temporal consistency and evolution significantly improved performance (AUC > 0.6 for all predictions), with disease progression alignment achieving the highest AUCs. These results suggest that pre-training the NODE layer with task-specific information is crucial for accurate DR progression prediction.

\begin{table}[]
\centering
\resizebox{0.89\columnwidth}{!}{%

%

\begin{tabular}{lllllllllll}
\hline
Method             & Weights & \multicolumn{3}{l}{Prediction 1 year} & \multicolumn{3}{l}{Prediction 2 year} & \multicolumn{3}{l}{Prediction 3 year} \\ \hline
                   &           & AUC1        & AUC2       & AUC3       & AUC1        & AUC2       & AUC3       & AUC1        & AUC2       & AUC3       \\ \hline
ResNet50           &    -      &  0.609      & 0.658      & 0.663      & 0.5747& 0.620 & 0.624   &  0.603 & 0.649 & 0.636           \\ 
ResNet50+NODE      &    -      & 0.533       & 0.624      & 0.568      & 0.538 & 0.572 & 0.599    & 0.501 & 0.503 & 0.512           \\
ResNet50+NODE-RNN  &    -      & 0.498       & 0.465      & 0.549      & 0.606 & 0.655 & 0.653   & 0.499 & 0.467 & 0.505          \\
ResNet50+NODE-LSTM &    -      & 0.524       & 0.619      & 0.564      & 0.504 & 0.459 & 0.575  & 0.542 & 0.616 & 0.615          \\
ResNet50+NODE-GRU  &    -      & 0.510       & 0.607      & 0.603     & 0.501 & 0.503 & 0.513     & 0.504 & 0.574 & 0.543             \\\hline

ResNet50+NODE      &    SimCLR-based     & 0.645       & 0.694      & 0.674      & 0.606 & 0.651 & 0.664     & 0.615 & \textbf{0.679} & \textbf{0.678}           \\
ResNet50+NODE-RNN  &    SimCLR-based     & 0.627       & 0.675      & 0.677      & 0.627 & 0.675 & 0.677     & 0.592 & 0.644 & 0.647           \\
ResNet50+NODE-LSTM &    SimCLR-based     & 0.617       & 0.669      & 0.672      & 0.622 & 0.669 & 0.665    & 0.603 & 0.653 & 0.630           \\
ResNet50+NODE-GRU  &    SimCLR-based     & 0.615       & 0.669      & 0.679     & 0.614 & 0.663 & 0.667     & 0.581 & 0.634 & 0.657           \\\hline

ResNet50+NODE      &    BYOL-based     & \textbf{0.654}       & \textbf{0.708}      & 0.669      & \textbf{0.654}& \textbf{0.708}& 0.669         & \textbf{0.640}& 0.685 & 0.673          \\
ResNet50+NODE-RNN  &    BYOL-based     & 0.650       & 0.696      & 0.680      & 0.650 & 0.696 & \textbf{0.680}& 0.608 & 0.659 & 0.661            \\
ResNet50+NODE-LSTM &    BYOL-based     & 0.622       & 0.669      & 0.665      & 0.640 & 0.679 & 0.625       & 0.609 & 0.654 & 0.667           \\
ResNet50+NODE-GRU  &    BYOL-based     & 0.654       & 0.697      & \textbf{0.698}& 0.605 & 0.667 & \textbf{0.680}& 0.612 & 0.654 & 0.673          
\end{tabular}
}
\caption{Comparison of method and pre-trained weights in terms of AUC for the prediction of diabetic retinopathy at 1, 2, and 3 years. Best results are in Bold.}
\label{tab:1_2_3_pred}
\end{table}

\noindent \textbf{Evaluation  for DR progression of time aware model on varying time interval (Task 2)}: Results shown in  Tab.\ref{tab:varible_pred} demonstrate the effectiveness of incorporating recurrent architectures (NODE-RNN, NODE-LSTM, NODE-GRU) and inspired pre-training strategies (SimCLR, BYOL) for predicting diabetic retinopathy progression. Compared to the baseline ResNet50+NODE model, all recurrent architectures achieved statistically significant improvements (p < 0.05, compared to the baseline without pretraining, DeLong test) in all three AUC metrics (AUC1, AUC2, AUC3) and in Kappa across all pre-training settings. Notably, BYOL pre-training consistently yielded the highest performance across all NODE-recurrent models, with the NODE-GRU variant achieving the best overall results (Kappa: 0.472, AUC1: 0.773, AUC2: 0.836, AUC3: 0.848).


\begin{table}[]
\begin{tabular}{cc}
    \begin{minipage}{.5\linewidth}

\centering
\resizebox{0.99\linewidth}{!}{%
\begin{tabular}{llllll}
\hline
Method             & Weights & \multicolumn{4}{l}{ \:\:\:\:\:\:\:\:\:\:\:\:\:\:\:\:\:\: Task 2} \\ \hline
                   &         & Kappa          & AUC1         & AUC2         & AUC3         \\
ResNet50+NODE      &     -   & 0.007 & 0.516 & 0.568 & 0.514              \\
ResNet50+NODE-RNN  &     -   & 0.042 & 0.561 & 0.623 & 0.628             \\
ResNet50+NODE-LSTM &    -   & 0.127 & 0.571 & 0.618 & 0.643            \\
ResNet50+NODE-GRU  &    -     & 0.105 & 0.571 & 0.639 & 0.672             \\\hline

ResNet50+NODE      &    SimCLR-based    & 0.090 & 0.560 & 0.631 & 0.680             \\
ResNet50+NODE-RNN  &    SimCLR-based    & 0.390 & 0.764 & 0.802 & 0.815             \\
ResNet50+NODE-LSTM &    SimCLR-based    & 0.459 & 0.763 & 0.825 & 0.800             \\
ResNet50+NODE-GRU  &    SimCLR-based    & 0.472 & 0.769 & \textbf{0.836} & 0.820             \\\hline

ResNet50+NODE      &     BYOL-based    & 0.177 & 0.589 & 0.646 & 0.714              \\
ResNet50+NODE-RNN  &    BYOL-based     & 0.421 & 0.760 & 0.831 & 0.847              \\
ResNet50+NODE-LSTM &    BYOL-based     & \textbf{0.511} & 0.764 & 0.830 & 0.841              \\
ResNet50+NODE-GRU  &    BYOL-based   & 0.459 & \textbf{0.773} & 0.835 & \textbf{0.848}   
\end{tabular}
}
\caption{Comparaison of method and pre-trained weights in term of AUC for the prediction of diabetic retinopathy for time varying interval. Best results are in Bold.}   
\label{tab:varible_pred}

\end{minipage} &

\begin{minipage}{.5\linewidth}

\resizebox{0.99\columnwidth}{!}{%
\begin{tabular}{llllll}\hline
Method                       & TC & Kappa & AUC1 & AUC2 & AUC3 \\ \hline
ResNet50+NODE (BYOL-based)   & No                  & 0.096 & 0.572 & 0.633 & 0.641      \\
ResNet50+NODE (BYOL-based)   & Yes                  & \textbf{0.177} & \textbf{0.589} &\textbf{ 0.646} & \textbf{0.714}       \\
\hline
                             & $\delta_{i}$         &  &  &  &  \\
ResNet50+NODE (SimCLR-based) &  1                     & 0.033 & 0.548 & 0.577 & 0.573     \\
ResNet50+NODE (SimCLR-based) &  2                  & 0.049 & 0.546 & 0.607 & 0.628    \\
ResNet50+NODE (SimCLR-based) &  3                  & \textbf{0.090} & \textbf{0.560} & \textbf{0.631} & \textbf{0.680}     
\end{tabular}
}
    \caption{Comparaison of the importance of configuration for BYOL/SimCLR-based pre-training technique on Task 2. Best results are in Bold.}
    \label{tab:tab_NODE_pretraining}
    \end{minipage} 
    
\end{tabular}
\end{table}

The findings reported in Tab.\ref{tab:1_2_3_pred}, \ref{tab:varible_pred} and \ref{tab:tab_NODE_pretraining} suggest that NODE-based recurrent architectures effectively capture temporal information in retinal disease progression. Pre-training with self-supervised learning methods inspired by SimCLR or BYOL further enhances model performance. One significant advantage of the proposed paradigm lies in its ability to train models with modular time steps. This flexibility contrasts with current methods, which often require imposing strict criteria on datasets to fit a specific prediction task. Such constraints can limit the model's ability to handle variable durations and sample numbers within the training data. While the tasks in our experiments share similar characteristics, we observed a greater performance gain for the task presented in Tab.\ref{tab:varible_pred} compared to Tab.\ref{tab:1_2_3_pred}. We hypothesize that this difference stems from the varying number of training samples between the two tasks. The results in Tab.\ref{tab:varible_pred} showcase the ability of our framework to fully leverage the strengths of continuous-time models, allowing us to train models with samples featuring diverse progression times (as depicted in Fig.\ref{fig:overview}.c) compared to the limitations of classical paradigms.

\subsection{Ablation study}

\noindent \textbf{Impact of value of the chosen $\delta_{i}$ and inverse constraint.} 
To examine the impact of temporal enhancement on model performance, we conducted an ablation study focusing on the values of $\delta_{i}$: (1) fixed: a constant value was assigned to $\delta_{i}$ for all data points; (2) variable (unaligned): $\delta_{i}$ was varied randomly for each data point, but not aligned with the progression of the disease; (3) variable (aligned): $\delta_{i}$ was varied based on the disease stage of each data point, aligned with the temporal dynamics of the disease as explained in Sec.\ref{secsub:BYOLinspired}. Tab.\ref{tab:tab_NODE_pretraining} highlights the crucial role of both the inverse constraint in the BYOL-inspired approach, and the alignment with disease progression and diverse values of $\delta_{i}$ in the SimCLR-inspired training strategy. 

\noindent \textbf{Convergence challenges and benefits of pre-training.} During experiments, when training our time-aware classifier using the "dopri5" solver, we encountered convergence issues, specifically when NODE was not pre-trained. These issues manifested as unstable learning curves and gradient instability during training.
Notably, pre-trained models exhibited significantly faster convergence and consistently stable gradients throughout the training process. This finding suggests that pre-training can effectively mitigate a prominent challenge associated with NODE training stability \cite{STEER}, aligning with previous observations reported in \cite{zeghlache-prime,zeghlache2023lmt,gong2021incorporating} about NODE pre-training and embedding capacities \cite{kuehn2023embedding}.

\section{Conclusion}

This work investigated pre-training NODEs, a type of time-aware model, for improved representation learning in the context of longitudinal medical image analysis. We achieved a significant improvement in performance on downstream longitudinal tasks focusing on predicting DR progression. Our novel time augmentation paradigm bridges the gap between time-aware methods and popular SSL approaches (demonstrated with BYOL/SimCLR), potentially generalizable to other techniques. This opens exciting avenues for exploring the re-training of TAM, particularly NODEs, for even more effective temporal modeling in medical imaging. Further investigations are warranted to explore the optimal hyperparameter settings for our framework and extend our contributions beyond classification tasks to encompass a wider range of tasks in a longitudinal setting.
\\

\noindent\textbf{Acknowledgments}
This work was conducted within the framework of the ANR RHU project Evired (ANR-18-RHUS-0008).


%
%
%
\bibliographystyle{splncs04}
\bibliography{biblio}

\end{document}


\title{
Supplementary for Longitudinal representation learning in continuous-time models to predict disease progression}

\author{***** \inst{1,2}
\and
***** \inst{1,3} 
\and
*****  \inst{1,2}
\and
*****  \inst{1,2}
\and
***** \inst{5}
\and
***** \inst{5} 
\and
***** \inst{5} 
\and
***** \inst{1,2,4} 
\and
***** \inst{1,2}
\and
*****  \inst{1,2} 
\and 
***** \inst{1} 
\and
***** \inst{1,2} 
}

\authorrunning{***** }
%

\institute{
*********** \and
*********** \and
***********
\and
***********\and
***********
}

\maketitle










\begin{table}[]
\centering
\begin{tabular}{|l|l|l|}
\hline
Method                        & Name               & Value                       \\ \hline
\multirow{4}{*}{BYOL based}   & Batch size         & 128                         \\
                              & MLP head dim       & 512                         \\
                              & Image size         & 256$\times$256              \\
                              & Neural ODE dim     & 512                         \\
                              & $\alpha$           & 0.99                        \\ \hline
\multirow{6}{*}{SimCLR based} & Batch size         & 128                         \\
                              & MLP head dim       & 512                         \\
                              & Neural ODE dim     & 512                         \\
                              & Fixed $\delta_{i}$ & 0,0125 (2 month equivalent) \\
                              & Image size         & 256$\times$256              \\
                              & $\tau$             & 0.5                         \\ \hline
\multirow{4}{*}{Fine-tuning}  & Image size         & 256                         \\
                              & Batch size         & 64                          \\
                              & Learning rate      & 0.001,0.0001,0.00001
                              \\
                              & Weight decay       & 0.0001                      \\ \hline
\end{tabular}
\caption{ Presentation of the hyperparameter settings used for BYOL and SimCLR based NODE pre-training methods, along with the fine-tuning scenarios applied for both Task 1 and Task 2.}
\end{table}

\begin{figure}[h!]
\centering
\includegraphics[width=1\textwidth]{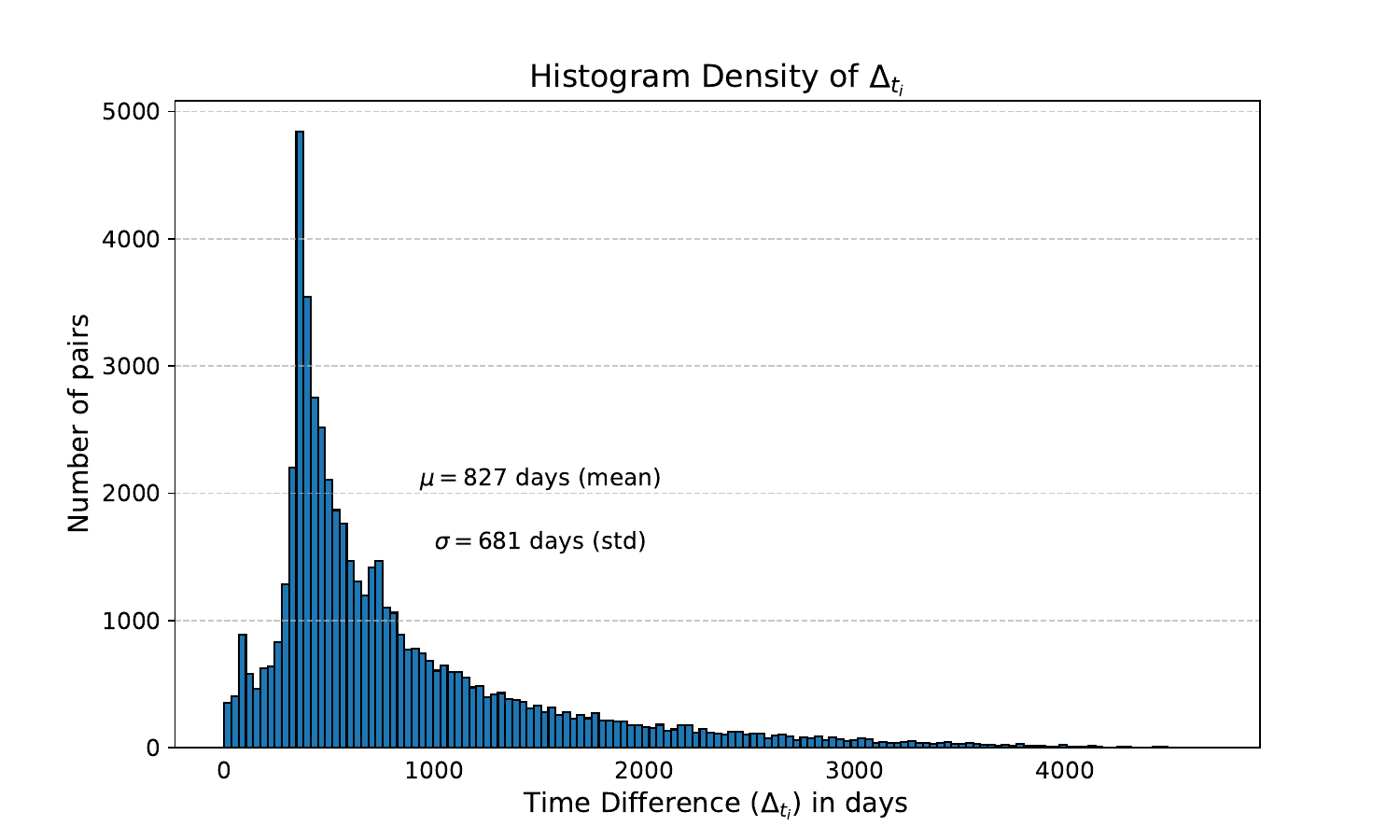}
\caption{Distribution of time differences ($\Delta_{t_{i}}$) across the datasets used for Task 2 and pre-training. This histogram shows the frequency of each time difference value. The mean and standard deviation of $\Delta_{t_{i}}$ are also displayed.} 

\label{fig:BYOL_SimCLR}
\end{figure}

\begin{table}[]
\center
\resizebox{1.3\columnwidth}{!}{%
\begin{tabular}{l|lll|l|}
\cline{2-5}
                             & \multicolumn{3}{l|}{Task 1: Prediction  of DR progression at an horizon of 3 years}                 & Task 2:  Prediction of DR at next available visits \\ \cline{2-5} 
                             & \multicolumn{1}{l|}{1 year prediction} & \multicolumn{1}{l|}{2 year prediction} & 3 year prediction & Prediction at variable time                        \\ \hline
No apparent DR               & \multicolumn{1}{l|}{62448}             & \multicolumn{1}{l|}{61153}             & 60218             & 25184                                              \\
Mild nonproliferative DR     & \multicolumn{1}{l|}{10311}             & \multicolumn{1}{l|}{11380}             & 11880             & 23928                                              \\
Moderate nonproliferative DR & \multicolumn{1}{l|}{3577}              & \multicolumn{1}{l|}{3792}              & 4205              & 7770                                               \\
Severe NPDR                  & \multicolumn{1}{l|}{645}               & \multicolumn{1}{l|}{655}               & 677               & 1056                                               \\
Proliferative DR (PDR)        & \multicolumn{1}{l|}{77}                & \multicolumn{1}{l|}{78}                & 78                & 128                                               
\end{tabular}
}
\caption{This table details the distribution of samples and classes for Tasks 1 and 2. The data originates from the ******* dataset, with two distinct subdatasets being used for analysis.} 
\end{table}